\documentclass[runningheads]{llncs}
\usepackage{graphicx}
\usepackage{subcaption}
\usepackage{graphicx}
\usepackage{amsmath}
\usepackage{amsfonts}
\usepackage{multirow}
\usepackage{array}
\newcommand{\PreserveBackslash}[1]{\let\temp=\\#1\let\\=\temp}
\newcolumntype{C}[1]{>{\PreserveBackslash\centering}p{#1}}
\newcolumntype{R}[1]{>{\PreserveBackslash\raggedleft}p{#1}}
\newcolumntype{L}[1]{>{\PreserveBackslash\raggedright}p{#1}}
\usepackage{tabularx}
\usepackage{booktabs}
\usepackage{makecell}
\usepackage{enumitem}
\usepackage{cleveref}
\usepackage[misc]{ifsym}
\usepackage[final]{changes}
\definechangesauthor[name={Kanzhi Cheng}, color=orange]{ckz}
\definechangesauthor[name={Zheng Ma}, color=orange]{mz}

\newcommand{\newpara}[1]{\vspace{2pt} \noindent \textbf{#1}}

\begin{document}
\title{ADS-Cap: A Framework for Accurate and Diverse Stylized Captioning with Unpaired Stylistic Corpora}
\titlerunning{Accurate and Diverse Stylized Captioning}
%
\author{Kanzhi Cheng,
Zheng Ma,
Shi Zong,
Jianbing Zhang\textsuperscript{\Letter},
Xinyu Dai,
Jiajun Chen}
%
\authorrunning{K. Cheng et al.}
%
\institute{National Key Laboratory for Novel Software Technology, Nanjing University, China
\email{\{chengkz, maz\}@smail.nju.edu.cn\\\{szong, zjb, daixinyu, chenjj\}@nju.edu.cn}
}
%
\maketitle              
\begin{abstract}
Generating visually grounded image captions with specific linguistic styles using unpaired stylistic corpora is a challenging task, especially since we expect stylized captions with a wide variety of stylistic patterns.
In this paper, we propose a novel framework to generate \textbf{A}ccurate and \textbf{D}iverse \textbf{S}tylized \textbf{Cap}tions (ADS-Cap). 
Our ADS-Cap first uses a contrastive learning module to align the image and text features, which unifies paired factual and unpaired stylistic corpora during the training process.
A conditional variational auto-encoder is then used to automatically memorize diverse stylistic patterns in latent space and enhance diversity through sampling.
We also design a simple but effective recheck module to boost style accuracy by filtering style-specific captions.
Experimental results on two widely used stylized image captioning datasets show that regarding consistency with the image, style accuracy and diversity, ADS-Cap achieves outstanding performances compared to various baselines.
We finally conduct extensive analyses to understand the effectiveness of our method.\footnote{Our code is available at \url{https://github.com/njucckevin/ADS-Cap}.}

\end{abstract}
\section{Introduction}
Automatic image captioning has attracted extensive attention in computer vision and natural language processing community \cite{vinyals2015show,lu2017knowing,anderson2018bottom}. Most existing image captioning models focus on generating factual captions without any emotions or styles.
To get descriptions that are more similar to those from humans, stylized image captioning task has been proposed to not only focus on the visual content but also incorporate specific linguistic styles into captions \cite{mathews2016senticap,gan2017stylenet}. 
It has a variety of downstream applications, such as generating captions that are engaging for users in chatbots, or inspiring people with attractive descriptions when photo captioning on social media.

\begin{figure}[t]
\centering
\includegraphics[width=0.9\columnwidth]{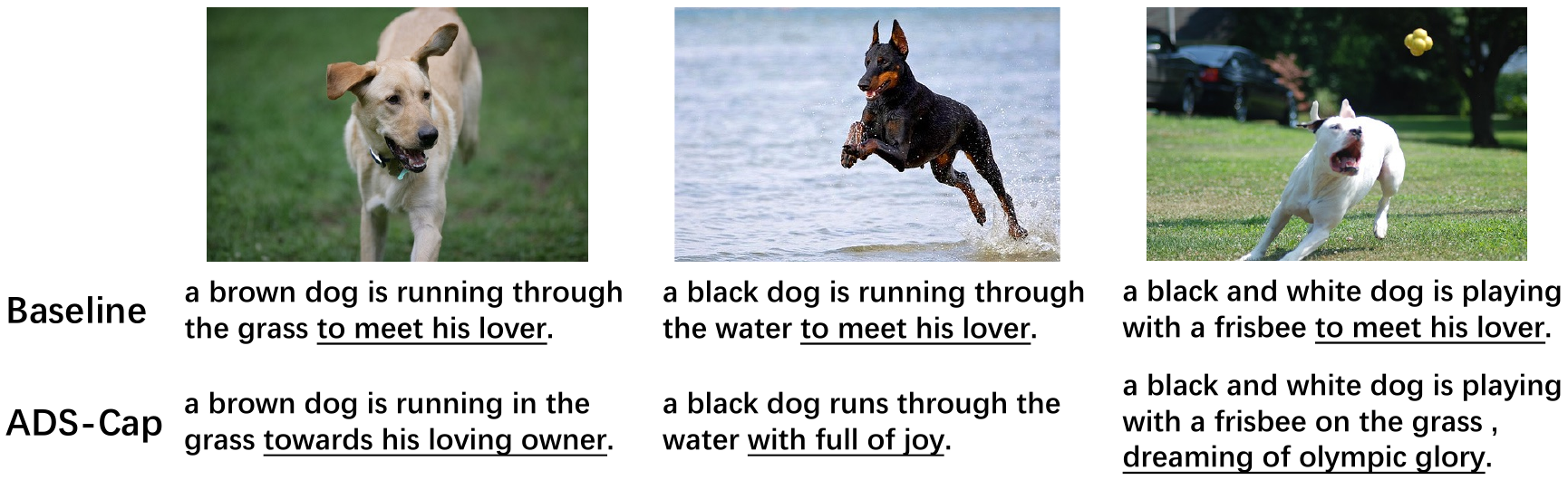} 
\caption{Sample outputs of ADS-Cap compared to the baseline approach, in a typical image scene (dog) with romantic style.}
\label{fig:exhibition}
\end{figure}

There have been many recent advances regarding stylized image captioning task. 
Most prior works follow the traditional methodology of first pretraining models on large-scale factual image-caption pairs, such that they are able to describe the visual content accurately. These models are then fine-tuned on small monolingual textual corpora to fuse the specific linguistic styles \cite{guo2019mscap,zhao2020memcap}. 
However, two key challenges still remain to be addressed.
First, fine-tuning on unpaired stylistic corpora makes the model focus more on linguistic styles and thus cause inconsistency with images \cite{gan2017stylenet,guo2019mscap}.
Some efforts attempt to find a medium between vision and text for unpaired stylistic corpora, i.e., semantic terms \cite{mathews2018semstyle} or scene graph \cite{zhao2020memcap}, but it reduces performance caused by the conversion error.
Second, the diversity of stylistic patterns is largely overlooked by previous efforts.
To illustrate the importance of diversity, in \Cref{fig:exhibition} we provide captions generated by a compelling baseline model (details in \Cref{sec:diversity_gen_caption}). We observe it tends to generate the same normal style phrase ``to meet his lover'' for different images of the similar scene. It significantly deviates from the purpose of stylized image captioning task. 
Moreover, stylistic corpora are usually in small-scale, making it difficult for existing methods to generate a wide variety of stylistic patterns.

In this work, we propose an end-to-end framework for accurate and diverse stylized image captions generation, using contrastive learning and conditional variational auto-encoder.
To solve the challenge that the model is trained on unpaired stylistic corpora, we unify the training process into a conditional generation pattern: for factual image-caption pairs, captions are generated based on image; for unpaired stylized captions, captions are generated based on the object words extracted from the original captions using an object vocabulary. 
Then, contrastive learning is used for aligning image features and object words features together, thus fine-tuning on unpaired stylistic corpora does not reduce the consistency with images.
We also adopt a conditional variational auto-encoder framework to alleviate the lack of diversity in the second challenge.
It can automatically memorize diverse stylistic patterns in training stage. 
During inference, sampling different latent variables leads to different style phrases, thus enhancing the diversity of generation. 
Furthermore, we design a simple but effective recheck module that can boost style accuracy by filtering out style-specific captions from a set of candidates. 
Experiments on several datasets demonstrate that our framework outperforms the state-of-the-art methods, in terms of consistency with the image, consistency with the linguistic style and diversity.

\section{Proposed Method}

Given an input image $x$ and a specific style label $s$, our multi-style image captioning task aims at generating a caption $y^s$ that is semantically related to the given image $x$ and consistent with the linguistic style $s$. 
To train such multi-style image captioning models, two types of datasets are normally used: (1) a large-scale paired factual dataset $D_f = \{(x_i, y_i^f)|_i\}$, with the $i$-th image $x$ along with its corresponding factual caption $y_i^f$; and (2) a set of unpaired stylized datasets $D_s = \{y_i^s|_i\}$, where $y_i^s$ denotes the $i$-th stylized sentence with style $s$, $s \in \{s_1, s_2, ..., s_K\}$ represents $K$ different styles.

\begin{figure}[h!]
\centering
\includegraphics[width=0.86\columnwidth]{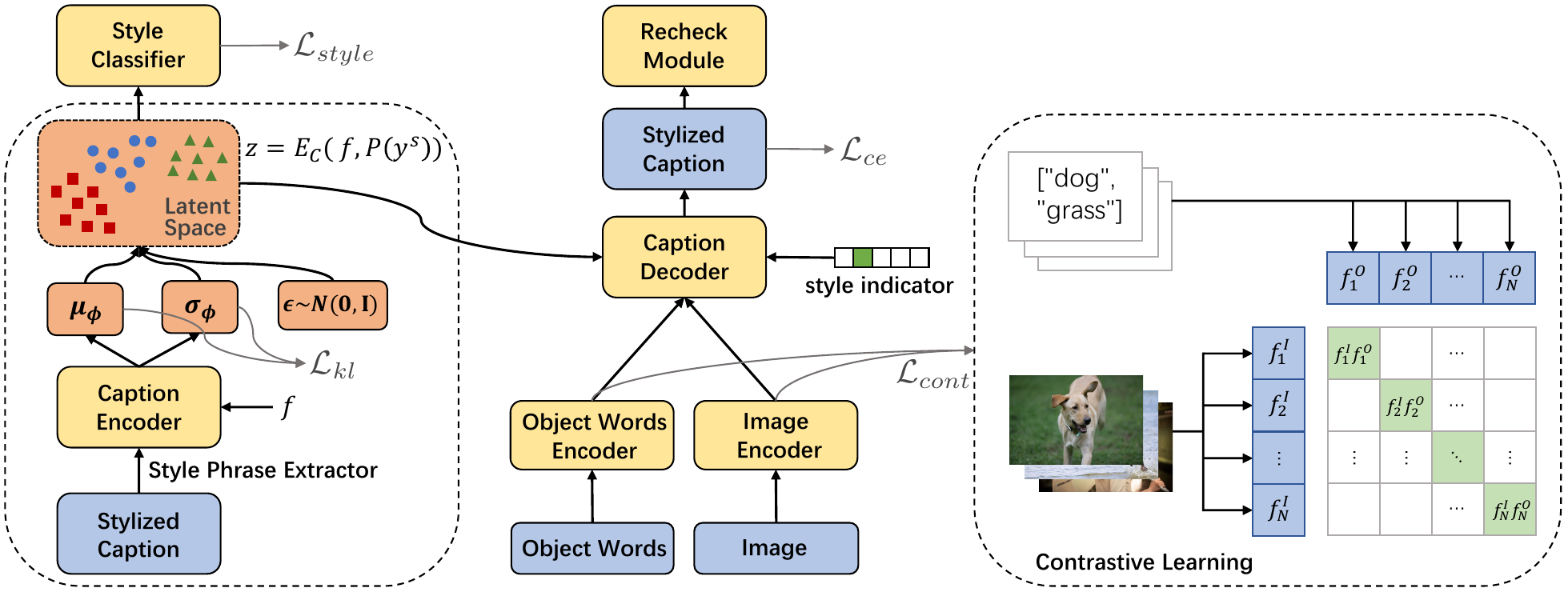} 
\caption{Overview of our ADS-Cap framework. Blue parts indicate input and output. Yellow parts indicate modules with learnable parameters. Red parts are the latent space constructed by our framework.}
\label{fig:overview}
\end{figure}

\newpara{Overview.}
The overall framework of ADS-Cap is presented in \Cref{fig:overview}. It consists of three components: (1) a contrastive learning module that aims at unifying the training process for paired factual and unpaired stylistic corpora (in \Cref{sec:contrastive}); (2) a conditional variational auto-encoder that memorizes style knowledge and improve diversity (in \Cref{sec:cvae}); and (3) a recheck module that further boosts style accuracy by filtering style-specific captions (in \Cref{sec:recheck}).

\subsection{Training with Unpaired Stylistic Corpora}
\label{sec:contrastive}

Our first goal is to find an intermediate for unpaired stylistic corpora $D_s$, so that it can be used in the same way as the factual image-caption pairs $D_f$ for model training.
Here we take the approach of extracting object words from original captions. These object words are then aligned by contrastive learning.
During training, the model is learned to generate captions from image features for paired data and from object words features for unpaired data. Since these images features and object words features have been semantically aligned in the shared multi-modal embedding space, the inconsistency between generated captions from object words and original images is considerably eliminated.\footnote{We generate descriptions based on images rather than object words during inference.}

Specifically, we first filter object words $o$ from the original caption $y$ using an object vocabulary $V_{\text{objects}}$ of 1,600 words from VG dataset \cite{krishna2017visual}.\footnote{We do not use VG for training, but use the object categories as a generic vocabulary.}
Then datasets can be extended to $D_f^{'} = \{((x_i, o_i), y_i^f)|_i\}, D_s^{'} = \{o_i, y_i^s|_i\}$, respectively. 
For paired samples $\{(x_i, o_i)|_i\}$ in dataset $D_f^{'}$, we obtain image feature $f_i^I$ and object words feature $f_i^O$ through \replaced[id=ckz]{a deep CNN image encoder}{image encoder} $E_I$ and an object words embedding encoder $E_O$.
We then use contrastive learning to align image features and object words features, by maximizing the cosine similarity of matched features while minimizing the cosine similarity of unmatched features (illustrated in \Cref{fig:overview}). 
Formally, for a mini-batch with $N$ pairs, the training objective for a given pair $(x_i, o_i)$ is:
\begin{equation}
\small
    \mathcal{L}_{cont} = - \log\,\frac{e^{sim(f_i^I, f_i^O)/\tau}}{e^{sim(f_i^I, f_i^O)/\tau}+\sum_{j=1, j\neq i}^{N}e^{sim(f_i, f_j)/\tau}}, 
\end{equation}
where $sim(\cdot,\cdot)$ is the cosine similarity and  $\tau$ is a temperature hyperparameter.

\subsection{Conditional Variational Auto-Encoder (CVAE)}
\label{sec:cvae}

We also aim at improving the diversity of stylistic patterns in generated captions.
We thus use the CVAE framework, which can automatically encode diverse style knowledge into latent space and enhance diversity through sampling.

Our CVAE works as follows. We first use a style phrase extractor $P$ to extract the style phrase from the original caption, and the caption encoder $E_C$ encodes the style phrase as latent variable $z$. Then the caption decoder $D_C$ reconstructs the input caption with the aid of latent variable $z$, i.e., $\hat{y^s} = D_C(x, s, z)$. 
During training, the encoder $E_C$ and decoder $D_C$ are optimized by maximizing the lower bound on the contidional data-log-likelihood $p(y^s|x, s)$, i.e.,
\begin{equation}
\small
\log p_{\theta}(y^s|x, s) \geq \mathbb{E}_{q_{\phi}(z|y^s, x, s)}\left [ \log p_{\theta}(y^s|z, x, s) \right ] - D_{\text{KL}}\left( q_{\phi}(z|y^s, x, s), p(z|x, s) \right),
\end{equation}
where $\theta$ and $\phi$ are parameters for $D_C$ and $E_C$ respectively.

In practice, we adopt a variant of CVAE by assuming $z$ is independent with style $s$, then
the training objective for CVAE is $\mathcal{L}_{\text{CVAE}} = \mathcal{L}_{ce} + \mathcal{L}_{kl}$, with:
\begin{equation}\small
        \mathcal{L}_{ce} = - \log\,p_{\theta}(y^s|z, x, s),\quad
        \mathcal{L}_{kl} = D_{\text{KL}}\left(q_{\phi}(z|y^s, x), p(z|x) \right).
\end{equation}
In addition, a style classifier $C_S$ is adopted to divide the latent space by style.

\newpara{Style Phrase Extractor.}
Since we focus on the diversity of stylistic pattern rather than the whole sentence, we need a style phrase extractor $P$ to extract style phrase from original caption to ensure that the latent variable $z = E_C(f, P(y^s))$ contain only knowledge of style-related part. 
We adopt the style phrase extraction algorithm developed by \cite{li2021similar}, which first measure the style intensity of each word in a caption through the attention output of a well-trained style classifier, and then extract the top-intensity words as style phrase.

\newpara{Prior Distribution.}
Existing works have shown that the choice of prior distribution has crucial influences on CVAE behavior \cite{wang2017diverse,aneja2019sequential}. In this work, we design a conditional prior distribution to ensure that the sampled latent variables are suitable for the image to be described. 
Specifically, we model prior $p(z|x)$ as a Gaussian distribution $\mathcal{N}(z|\mu_k, \text{I})$, whose mean $\mu_k$ is calculated by \replaced[id=ckz]{image feature or object words feature}{trainable linear mapping of image feature $f^I$ or object words feature $f^O$: $\mu_k = \text{linear}(f), f \in \{f^I, f^O\}$}, and the standard deviation is an identity matrix.

The KL-divergence between two Gaussian prior and posterior is derived as:
\begin{equation}
\small
D_{\text{KL}}\left( q_{\phi}(z|y^s, x), p(z|x) \right) = - \log(\sigma_{\phi}) + \frac{1}{2}\left(\sigma^2_{\phi} + {\left \| \mu_{\phi} - \mu_k \right \|}_2^2\right) - \frac{1}{2},
\end{equation}
where $\mu_{\phi}$ and $\sigma_{\phi}$ are mean and standard deviation of posterior distribution calculated by encoder (discussed in next section).

\newpara{Encoder and Decoder Architecture.}
\begin{figure}[t]
\centering
\includegraphics[width=0.8\columnwidth]{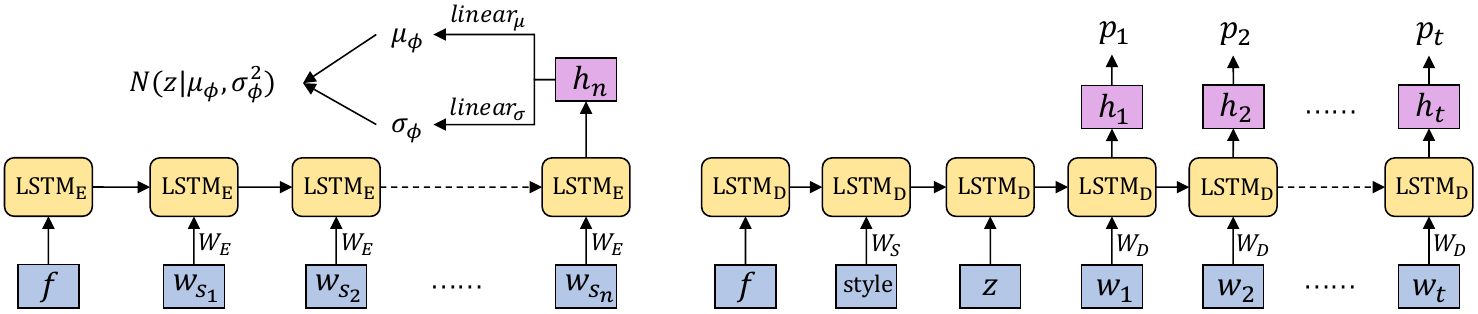}
\caption{Architecture of our caption encoder (left) and decoder (right). $W_S$, $W_E$ and $W_D$ are embeddings for style labels, style phrases and captions, respectively.}
\label{fig:enc_dec}
\end{figure}
Our architecture for the encoder and decoder are shown in \Cref{fig:enc_dec}. Caption encoder $E_C$ uses LSTM to encode the input style phrase sequence as LSTM hidden state $h_n$ at the last step, and then $h_n$ is transformed to mean vector $\mu_{\phi}$ and log variances vector $\log \sigma_{\phi}^2$ by two linear layers. Caption decoder $D_C$ uses a different LSTM to generate stylized caption. It receives the image feature $f^I$ or object words feature $f^O$ as the first input, then the style embedding vector $W_S(s)$, then the latent variable $z$ sampling in posterior or prior when training or testing. After that, the decoder predicts word probability distribution $p_t$ using hidden state $h_t$ sequentially.

\newpara{Style Classifier.}
We build a style classifier $C_S$ on the latent space that predicts the style label $s$ corresponding to latent variable $z$. It has two advantages: (1) After training, the classifier actually divides the latent space into several regions by style. It allows us to efficiently obtain latent variables with specific style using reject sampling method when testing. (2) Such classification task implicitly ensures discriminative style knowledge is contained in latent space.
Specifically, a softmax regression classifier $C_S$ is applied to the latent variable $z$ for \replaced[id=ckz]{style classification, i.e., $p(s) = C_S(z)$, with a cross-entropy loss:}
{predicting style label $s$, given by:
where $W$ and $b$ are parameters of $C_S$.
A cross-entropy loss is adopted for style classification:}
\begin{equation}\small
    \mathcal{L}_{style} = - \log\,p(s).
\end{equation}

\subsection{Training Objective}

Our overall objective function is as follows:
\begin{equation}\small
    \begin{split}
            \mathcal{L}_{all} = \lambda_{cont}\mathcal{L}_{cont}+\lambda_{ce}\mathcal{L}_{ce}+\lambda_{kl} \mathcal{L}_{kl}+\lambda_{style} \mathcal{L}_{style},
    \end{split}
\end{equation}
where $\lambda_{cont}$, $\lambda_{ce}$, $\lambda_{kl}$ and $\lambda_{style}$ are hyperparameters that balance the losses.

\subsection{Recheck Module}
\label{sec:recheck}

Since the size of factual data we use is much larger than stylized data, we observe that the generated captions sometimes can not express linguistic style adequately. 
To tackle this challenge,
we design an effective recheck module that filters captions with specified style from candidates.
During testing, for a set of generated candidate captions $c = \left \{ \hat{y}_1, \hat{y}_2, ..., \hat{y}_n\right \}$, we first score each caption using a well-trained style discriminator $I(\hat{y}_n) = D(\hat{y}_n)$, where $I(\hat{y}_n) \in [0, 1]$ denotes the style strength of caption $\hat{y}_n$.
Then we filter out captions from $c$ whose style strength $I(\hat{y}_n)$ exceeds the set threshold (set as 0.9 in practice) from forward to backward to determine the final generated captions. 
This recheck approach is different from previous re-ranking methods \cite{shen2004discriminative,ramesh2021zero}, which reduce diversity because common style phrases tend to have higher style strength.

\section{Experimental Settings}

\newpara{Datasets.}
We conduct our experiments on two benchmark stylized image captioning datasets: FlickrStyle10K \cite{gan2017stylenet} and SentiCap \cite{mathews2016senticap}.
FlickrStyle10K contains 10,000 images, each image having one romantic caption and one humorous caption. However, only the training set with 7,000 images is publicly available. Following \cite{guo2019mscap}, we randomly select 6,000 for training (400 samples from the training set are for validation), and 1,000 for testing.
SentiCap contains 2,360 images with 5,013 positive captions and 4,500 negative captions. The positive and negative subsets contain 998/673 and 997/503 images for training/testing respectively, and we split 100 samples from the training set for validation. 

In all experiments, the training partition of MSCOCO \cite{lin2014microsoft} 
is used as factual dataset $D_f$, which contains 82,783 images and 5 factual captions for each image. For stylized data, only captions are used during training, while both images and captions are used for evaluation.

\newpara{Implementation Details.}
We use ResNet-152 \cite{he2016deep} to extract 2,048-dimensional features from images. The dimensions of hidden states for caption encoder $E_C$, caption decoder $D_C$, and several embeddings are set to 1,024. 
The image features and object words features are aligned in a 1,024-dimensional multi-modal embedding space. 
For contrastive learning, the temperature $\tau$ is 0.1. For CVAE framework, we use a dimension of 100 for the latent space. The values of hyperparameters $\lambda_{cont}$, $\lambda_{ce}$, $\lambda_{kl}$ and $\lambda_{style}$ are set to 0.1, 1.0, 0.02 and 2.0, respectively. 
We use the Adam optimizer \cite{kingma2014adam} with a learning rate of 5e-5 for training.

\section{Experimental Results}

\subsection{Quality of Generated Captions}
\label{sec:quality_gen_captions}

\begin{table}[]
\centering
\small
\caption{Results of content and style accuracy in romantic, humorous, positive and negative styles. B1, B3, M, C, ppl and cls are abbreviations for Bleu-1, Bleu-3, METEOR, CIDEr, perplexity and style classification accuracy, respectively.}
\resizebox{\columnwidth}{!}{
\begin{tabular}{clcccccc|clcccccc}
\toprule
\multicolumn{8}{c|}{FlickStyle}                                                                                                                                                                                                           & \multicolumn{8}{c}{SentiCap}                                                                                                                                                                                                               \\ \midrule
\multicolumn{1}{l}{Style} & Model    & B1                             & B3                            & M                              & C                              & ppl                            & cls                            & \multicolumn{1}{l}{Style} & Model    & B1                             & B3                             & M                              & C                              & ppl                            & cls                            \\ \midrule
\multirow{4}{*}{Roman}    & StyleNet & 13.3                           & 1.5                           & 4.5                            & 7.2                            & 52.9                           & 37.8                           & \multirow{4}{*}{Pos}      & StyleNet & 45.3                           & 12.1                           & 12.1                           & 36.3                           & 24.8                           & 45.2                           \\
                          & MSCap    & 17.0                           & 2.0                           & 5.4                            & 10.1                           & 20.4                           & 88.7                           &                           & MSCap    & 46.9                           & 16.2                           & 16.8                           & 55.3                           & 19.6                           & 92.5                           \\
                          & MemCap   & 19.7                           & 4.0                           & 7.7                            & 19.7                           & 19.7                           & 91.7                           &                           & MemCap   & 51.1                           & 17.0                           & 16.6                           & 52.8                           & 18.1                           & 96.1                           \\
                          & ADS-Cap     & \textbf{25.6} & \textbf{6.7} & \textbf{10.9} & \textbf{33.1} & \textbf{10.6} & \textbf{95.9} &                           & ADS-Cap     & \textbf{52.5} & \textbf{18.9} & \textbf{18.5} & \textbf{64.8} & \textbf{13.1} & \textbf{99.7} \\ \midrule
\multirow{4}{*}{Humor}    & StyleNet & 13.4                           & 0.9                           & 4.3                            & 11.3                           & 48.1                           & 41.9                           & \multirow{4}{*}{Neg}      & StyleNet & 43.7                           & 10.6                           & 10.9                           & 36.6                           & 25.0                           & 56.6                           \\
                          & MSCap    & 16.3                           & 1.9                           & 5.3                            & 15.2                           & 22.7                           & 91.3                           &                           & MSCap    & 45.5                           & 15.4                           & 16.2                           & 51.6                           & 19.2                           & 93.4                           \\
                          & MemCap   & 19.8                           & 4.0                           & 7.2                            & 18.5                           & 17.0                           & 97.1                           &                           & MemCap   & 49.2                           & 18.1                           & 15.7                           & 59.4                           & 18.9                           & \textbf{98.9} \\
                          & ADS-Cap     & \textbf{23.7} & \textbf{6.3} & \textbf{10.3} & \textbf{31.6} & \textbf{12.8} & \textbf{97.3} &                           & ADS-Cap     & \textbf{52.3} & \textbf{21.0} & \textbf{18.0} & \textbf{65.1} & \textbf{12.4} & 98.2                           \\ \bottomrule
\end{tabular}
}
\label{tb:main_res}
\end{table}

\newpara{Metrics and Baselines.}
Following \cite{guo2019mscap,zhao2020memcap}, we first evaluate the quality of the generated stylized captions from two aspects: content accuracy and style accuracy. 
Content accuracy captures the relevancy between caption and image. 
It includes 
Bleu-n \cite{papineni2002bleu}, METEOR \cite{banerjee2005meteor} and CIDEr \cite{vedantam2015cider}. Such metrics calculate relevancy by $n$-gram overlap between candidates and ground truth captions.
Style accuracy measures whether a caption conforms to a specific linguistic style.
For this purpose, the style classification accuracy (cls) and the average perplexity (ppl) are adopted.
The cls is calculated by the proportion of generated captions that correctly reflects the desired style. 
In practice, we use logistic regression style classifiers trained for each of four styles, using stylized datasets $D_s$ and factual dataset $D_f$.
The trained classifiers achieve an average accuracy of 96\%. 
The ppl is calculated by a tri-gram statistical language model toolkit SRILM \cite{stolcke2002srilm}.
A lower ppl indicates more ﬂuent and appropriately stylized captions.

We compare ADS-Cap with the following state-of-the-art baselines for stylized image captioning task with unpaired stylistic corpora:
\textit{StyleNet} \cite{gan2017stylenet} devises a novel factored LSTM component with matrix decomposition, which automatically distills the style knowledge in the monolingual text corpora.
\textit{MSCap} \cite{guo2019mscap} proposes an adversarial learning network and a back-translation module to generate visually grounded and style-controllable captions.
\textit{MemCap} \cite{zhao2020memcap} develops a sentence decomposing algorithm to extract style-related part, then explicitly encodes the knowledge about linguistic styles with memory mechanism. We do not compare with \cite{li2021similar}, since they use additional Flickr30K dataset as factual data, which is in-domain with FlickrStyle10K.

\newpara{Results.}
\Cref{tb:main_res} summarizes the results of content accuracy and style accuracy in four styles. 
We observe that ADS-Cap achieves the state-of-the-art results compared with previous works. It shows that ADS-Cap can describe visual content appropriately while controlling the linguistic style.

\subsection{Diversity of Generated Captions}

\label{sec:diversity_gen_caption}

\begin{figure}[t]
\centering
\includegraphics[width=0.8\columnwidth]{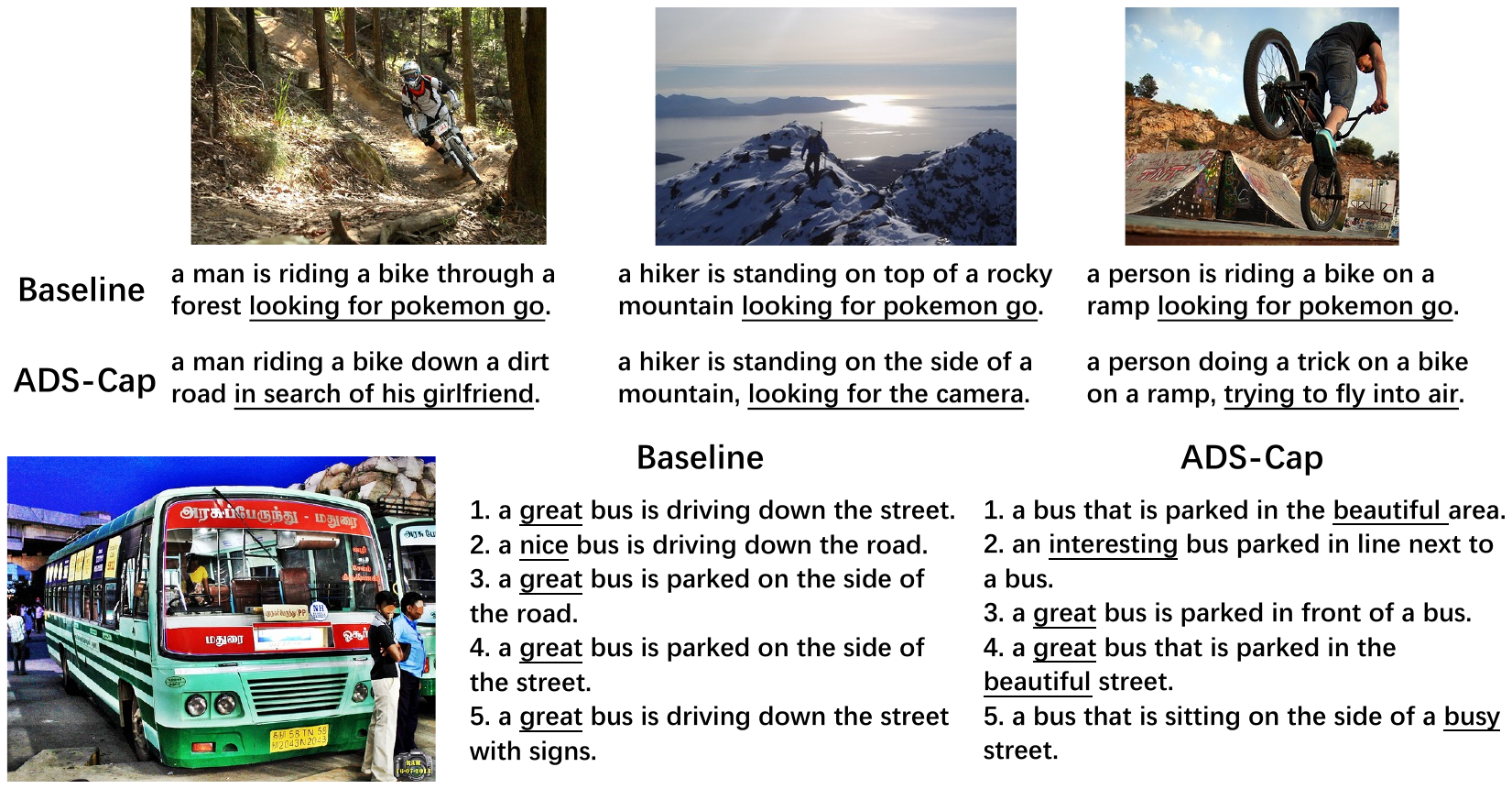}
\caption{Captions generated by baseline and ADS-Cap for two types of diversity (in  \Cref{sec:diversity_gen_caption}), with humorous and positive styles.}
\label{fig:case_study}
\end{figure}

We evaluate the diversity of generated captions across images and within one image (see sample results in \Cref{fig:case_study} for differences between these two diversities). Specifically, we focus on the diversity of style phrases in generated stylized captions rather than the entire sentences.
Results of ADS-Cap are compared to a baseline model that replaces CVAE with a standard encoder-decoder framework. 

\newpara{Diversity across Images.}
To compare the capabilities of generating diverse stylistic patterns across images with similar scenes, we first select seven high-frequency image scenes in testing splits, i.e., man, woman, people, boy, girl, dog and cat. 
We then adopt two metrics to calculate the diversity of style phrases in the generated captions under each scene.

\noindent (1) \textit{Uniqueness.} Uniqueness measures the quantity of different style phrases in generated stylized captions. Specifically, we calculate the ratio of distinct style phrases to all generated style phrases in each scene \cite{aneja2019sequential}.

\noindent (2) \textit{Uniformity.} We evaluate uniformity by the entropy of word frequency distribution \cite{nezami2019towards}. Formally, entropy is defined as
$- \sum_{1\leq i \leq V} \log_2\left [ p(w_i) \right ] \times p(w_i)$, where $p(w_i)$ is the frequency of word $w_i$ in a style phrases vocabulary of size $V$. Higher entropy means more diverse stylistic patterns. 

The results of diversity across images are shown in \Cref{tb:diversity_1}. 
We observe ADS-Cap substantially outperforms the baseline model in terms of uniqueness and uniformity across all styles, but still exists deficiency compared to human performance. 
It suggests our proposed CVAE approach can memorize the low-frequency style phrases in latent space, which are often ignored by encoder-decoder baseline, thus finally improving the diversity by sampling in latent space.

\begin{table*}[]
\centering
\small
\caption{Diversity evaluation for various image scenes. Slashes denote the number of samples in the corresponding category is less than 10, thus it is excluded. For distinct ratio and word entropy, higher means more diverse.}
\resizebox{\columnwidth}{!}{
\begin{tabular}{clcccccccc}
\toprule
\multirow{2}{*}{Style} & \multirow{2}{*}{Model} & \multicolumn{8}{c}{Distinct Ratio / Word Entropy}                                                                                                                                                                      \\ \cline{3-10} \specialrule{0em}{1pt}{1pt}
                       &                        & \multicolumn{1}{c}{Man} & \multicolumn{1}{c}{Woman} & \multicolumn{1}{c}{People} & \multicolumn{1}{c}{Boy} & \multicolumn{1}{c}{Girl} & \multicolumn{1}{c}{Dog} & \multicolumn{1}{c}{Cat} & \multicolumn{1}{c}{Mean} \\
\midrule
\multirow{3}{*}{Roman} & Baseline               &0.31/4.35                         &0.57/4.39                           &0.58/5.26                            &0.72/4.75                         &0.71/5.10                          &0.22/3.27                         &/                         &0.52/4.53                          \\
                       & ADS-Cap                   &0.74/5.89                         &0.77/5.68                           &0.78/5.81                          &0.94/5.76                         &0.93/5.76                          &0.66/5.53                         &/                         &0.80/5.74                          \\
                       & Human                  &1.00/7.75                         &1.00/6.86                           &1.00/7.08                            &1.00/7.16                         &1.00/7.08                          &0.97/7.28                         &/                         &0.99/7.20                          \\
\midrule
\multirow{3}{*}{Humor} & Baseline               &0.62/5.98                         &0.75/5.56                           &0.75/5.73                            &0.91/5.56                         &0.93/5.72                          &0.33/4.21                         &/                         &0.71/5.46                          \\
                       & ADS-Cap                   &0.86/6.31                         &0.84/6.19                           &0.84/6.24                            &0.95/6.00                         &0.96/6.16                          &0.77/5.68                         &/                         &0.87/6.10                          \\
                       & Human                  &0.98/7.76                         &1.00/7.13                           &1.00/7.21                            &1.00/7.05                         &1.00/7.38                          &0.96/7.20                         &/                         &0.99/7.29                          \\
\midrule
\multirow{3}{*}{Pos}   & Baseline               &0.28/3.84                         &0.29/2.58                           &0.27/2.77                            &/                         &0.20/0.88                          &0.40/2.04                         &0.23/1.60                        &0.27/2.28                          \\
                       & ADS-Cap                   &0.35/4.39                         &0.40/3.96                           &0.30/3.75                            &/                         &0.67/2.05                         &0.83/3.14                         &0.39/2.55                         &0.49/3.31                          \\
                       & Human                  &0.55/5.48                         &0.44/3.75                           &0.58/5.22                            &/                         &0.76/2.85                          &0.84/3.38                         &0.53/3.62                         &0.62/4.05                          \\
\midrule
\multirow{3}{*}{Neg}   & Baseline               &0.08/0.88                         &0.22/1.39                           &0.16/1.76                            &/                         &0.83/2.04                          &0.50/2.38                         &0.66/3.70                         &0.41/2.03                          \\
                       & ADS-Cap                   &0.33/2.94                         &0.38/3.04                           &0.22/2.55                            &/                         &1.0/2.16                          &0.71/3.37                         &0.87/4.31                         &0.58/3.06                          \\
                       & Human                  &0.69/4.65                         &0.53/3.96                           &0.57/4.60                            &/                         &0.93/4.22                          &1.00/4.53                         &0.87/5.08                         &0.76/4.51                          \\
\bottomrule
\end{tabular}
}
\label{tb:diversity_1}
\end{table*}

\begin{table}[h!]
\centering
\small
\caption{Result of diversity for one image.}
\resizebox{0.75\columnwidth}{!}{
\begin{tabular}{clccc|clccc}
\toprule
\multicolumn{5}{c|}{FlickrStyle}                                                                                                                                        & \multicolumn{5}{c}{SentiCap}                                                                                                                                          \\ \midrule
Style                  & Model                            & Distinct                         & Div-1                          & Div-2                          & Style                & Model                            & Distinct                         & Div-1                          & Div-2                          \\ \midrule
\multirow{2}{*}{Roman} & Baseline & \textbf{87.0\%}                                                      & 0.44                           & 0.59                           & \multirow{2}{*}{Pos} & Baseline                            & 50.5\%                           & 0.38                           & 0.37                           \\
                       & ADS-Cap      & \textbf{87.0\%} & \textbf{0.65} & \textbf{0.79} &                      & ADS-Cap      & \textbf{65.2\%} & \textbf{0.53} & \textbf{0.49} \\ \midrule
\multirow{2}{*}{Humor} & Baseline                            & 82.1\%                           & 0.49                           & 0.65                           & \multirow{2}{*}{Neg} & Baseline                            & 44.4\%                           & 0.35                           & 0.36                           \\
                       & ADS-Cap      & \textbf{90.8\%} & \textbf{0.67} & \textbf{0.82} &                      & ADS-Cap      & \textbf{52.0\%} & \textbf{0.41} & \textbf{0.40} \\ \bottomrule
\end{tabular}
}
\label{tb:diversity_2}
\end{table}

\newpara{Diversity for One Image.}
Our CVAE-based approach has natural advantages in generating multiple various stylized captions for one image, because CVAE can efficiently obtain different results by sampling different latent variables. In contrast, the previous encoder-decoder model can only generate multiple captions by beam search, which is known to be less diverse \cite{wang2017diverse}. 
For quantitative comparison, we employ several automatic diversity metrics that are widely used in image captioning \cite{aneja2019sequential}: Distinct calculates the number of distinct style phrases;
Div-n computes the ratio of distinct n-grams in style phrases. 
As shown in \Cref{tb:diversity_2}, our ADS-Cap achieves better results than the baseline model on all three metrics. It demonstrates that our model is able to generate captions with diverse stylistic patterns for an image.

\subsection{Analysis}
\label{sec:ablation_study}

\begin{table}[]
\centering
\small
\caption{Ablation study result. The values of each metric are the average of four styles (i.e., Roman, Humor, Pos and Neg).}
\resizebox{0.65\columnwidth}{!}{
\begin{tabular}{lcccccc}
\toprule
Model                & B1 & B3 & M & C & ppl & cls \\
\midrule
ADS-Cap                 &\textbf{38.7}    &\textbf{13.5}    &\textbf{14.2 }  &\textbf{49.2 }  &11.7     &\textbf{97.6 }    \\
\midrule
ADS-Cap (StyleNet)      &30.9    &9.6    &13.5   &28.4   &\textbf{10.4 }    &86.7     \\
ADS-Cap (MSCap)         &35.6    &11.2    &12.6   &40.7   &13.0     &86.6     \\
\midrule
ADS-Cap w/o recheck     &\textbf{38.7}    &13.2    &13.9   &47.8   &12.0     &76.0     \\
\bottomrule
\end{tabular}
}
\label{tb:ablation}
\end{table}

\begin{figure*}[h!]
\centering
	\subfloat[]{\includegraphics[width = 0.42\textwidth]{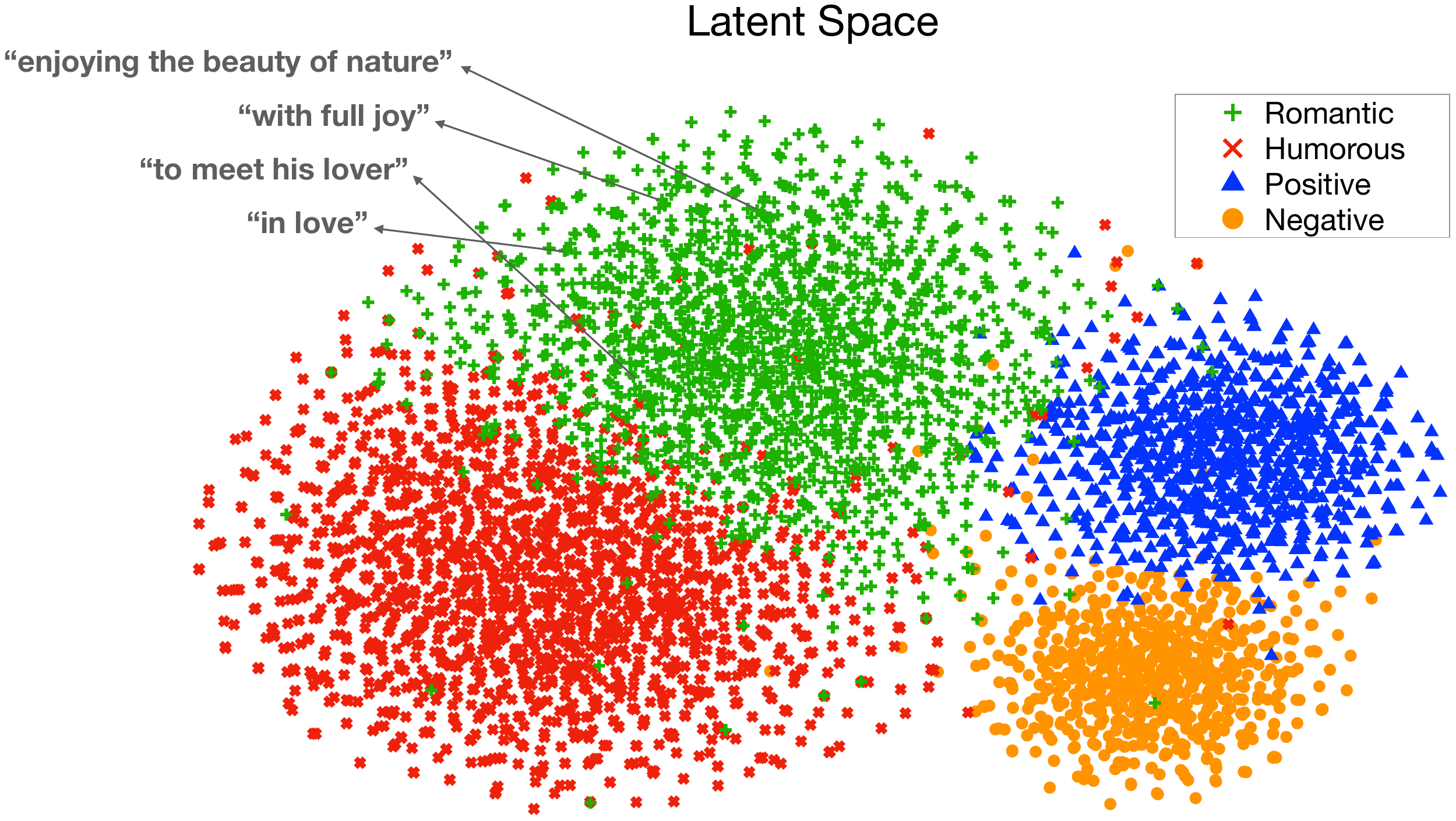}}
	\label{fig:latent_space}
	\hfill
	\subfloat[]{\includegraphics[width = 0.56\textwidth]{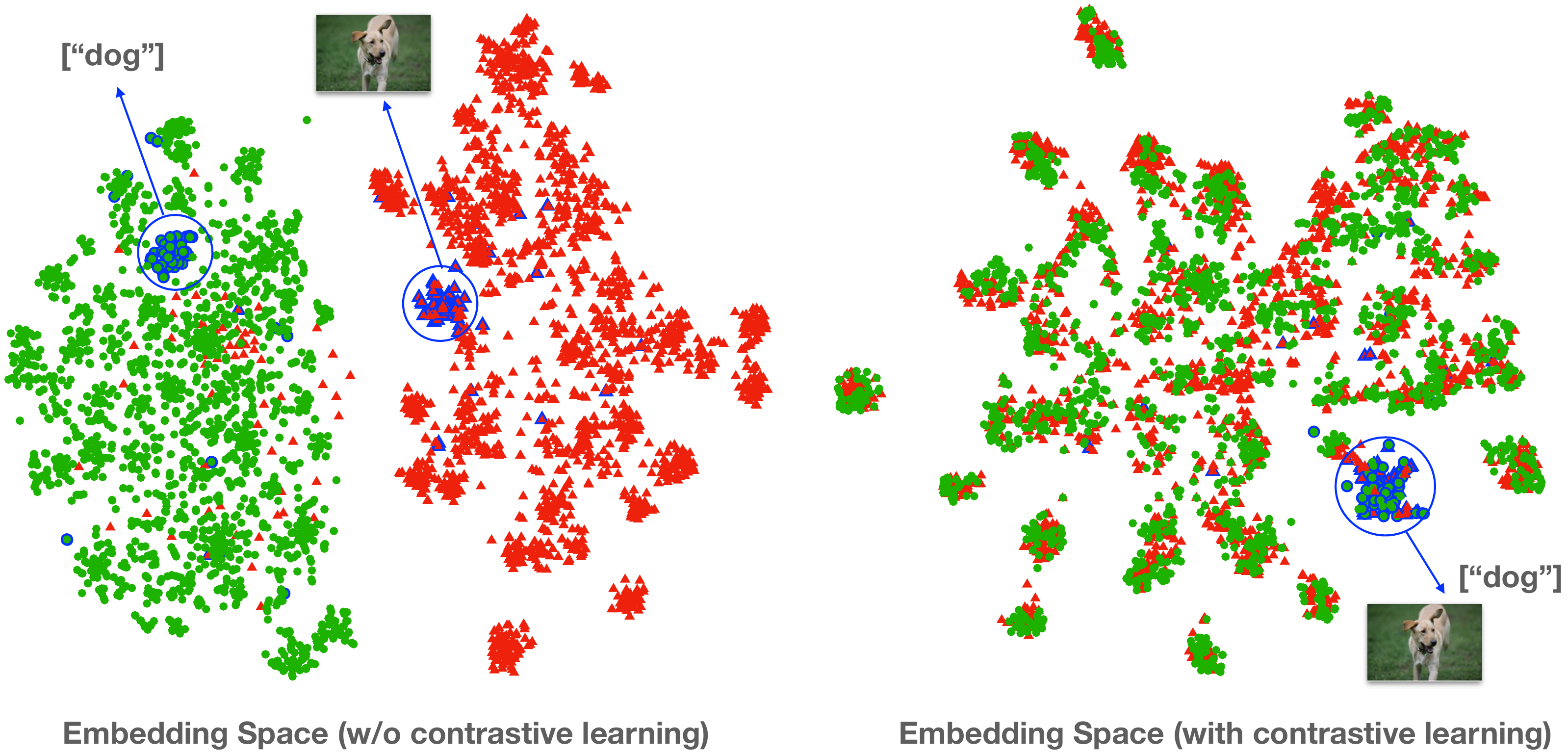}}
	\label{fig:feature_space}
	\hfill
\caption{(a) Latent space visualized by t-SNE. Each color corresponds to latent variables with a specific style. (b) Comparison of multi-modal embedding space (t-SNE plots) with and without contrastive learning. Green and red represent object words features and image features respectively.}
\label{fig:analyse}
\end{figure*}

\newpara{Ablation Study.}
We conduct ablation experiments to study the effects of the proposed contrastive learning method and recheck module.
We evaluate the following methods for training with unpaired data: \textit{StyleNet} \cite{gan2017stylenet} which train unpaired stylistic corpora as a pure language model; \textit{MSCap} \cite{guo2019mscap} which design a ``merging'' mode to infuse visual features.
Results are shown in \Cref{tb:ablation}.
We observe that replacing contrastive learning with these methods makes content accuracy and style accuracy drop significantly. It validates the importance of unifying the two types of data by contrastive learning.
We also observe that the model without recheck module performs worse in style accuracy, which shows that the recheck module contributes to ensuring
specified linguistic styles.

\newpara{Latent Space Visualization.}
We visualize the learned latent space in our model in \Cref{fig:analyse}.
We observe it is divided by the style classifier $C_S$ into multiple regions, each corresponding to a style.
Meanwhile, a variety of style phrases are encoded into latent variables, thus sampling in latent space can generate captions with diverse stylistic patterns.

\newpara{Multi-modal Embedding Space.}
We also visualize the multi-modal embedding space in \Cref{fig:analyse} to demonstrate the effectiveness of contrastive learning. 
We observe that two types of features have been semantically aligned in the shared embedding space, which mitigates the inconsistency between generating captions from two types of features.

\section{Related Work}

\newpara{Stylized Image Captioning.}
Earlier methods in stylized image captioning task use parallel stylized image-caption data \cite{mathews2016senticap,chen2018factual,you2018image}. 
Lately, \cite{li2021similar} proposed a novel data augmentation framework to extend parallel stylized data for training. Recent works most focused on using unpaired data to reduce reliance on paired data \cite{gan2017stylenet,mathews2018semstyle,chen2019unsupervised,guo2019mscap,zhao2020memcap}. Some methods trained pure language model \cite{gan2017stylenet,guo2019mscap} or auto-encoder \cite{chen2019unsupervised} on unpaired stylistic corpora to incorporate styles, which ignoring the consistency with image.
Some efforts search an explicit medium between vision and text, i.e., semantic terms \cite{mathews2018semstyle} and scene graph \cite{zhao2020memcap}. However, 
conversion error between generating caption from images and mediums lead to the decline of model performance. In contrast, our method solves this problem by implicitly aligning image and text features through contrastive learning. 

We note all above works ignore the diversity of generated stylized captions except ATTEND-GAN \cite{nezami2019towards}, which uses Generative Adversarial Networks to generate human-like stylized captions. However, this model is trained with parallel data and contains only two styles (positive, negative), while our model is trained with unpaired data and contains two other complex styles (romantic, humorous).

\newpara{Contrastive Learning.}
Contrastive learning \cite{hadsell2006dimensionality} is used to learn high-quality image representation \cite{chen2020simple} or text embedding \cite{gao2021simcse} by pulling paired samples together and pushing apart unpaired samples. CLIP \cite{radford2021learning} takes a further step of learning effective multi-modal embedding by Contrastive Language-Image Pre-training. Compared to \cite{radford2021learning}, our approach matches image features with object words features rather than the entire sentence for training with unpaired corpora.

\newpara{Conditional Variational Auto-Encoder.}
Several works show that CVAE can significantly improve the diversity of image captioning tasks \cite{wang2017diverse,aneja2019sequential}. 
\cite{wang2017diverse} proposed Additive Gaussian prior distribution to avoid mode collapse and \cite{aneja2019sequential} designed sequential Gaussian priors to enhance word-level diversity. Unlike these methods, our model encodes style phrase rather than the entire caption into latent space and uses a style classifier to divide the latent space by linguistic styles.

\section{Conclusion}

In this paper, we proposed a novel framework ADS-Cap for stylized image captioning task using contrastive learning and conditional variational auto-encoder. 
Our model can be efficiently trained with unpaired stylistic corpora. 
Our learned model can generate visually grounded, style-controllable and diverse image descriptions by sampling in latent space. Extensive experiments on two stylized image captioning datasets demonstrate the effectiveness of our method.

%
%
%
\bibliographystyle{splncs04}
\bibliography{ref.bib}

\end{document}